\documentclass[pdflatex,sn-mathphys-ay]{sn-jnl}

\usepackage{graphicx}%
\usepackage{multirow}%
\usepackage{amsmath,amssymb,amsfonts}%
\usepackage{amsthm}%
\usepackage{mathrsfs}%
\usepackage[title]{appendix}%
\usepackage{xcolor}%
\usepackage{textcomp}%
\usepackage{manyfoot}%
\usepackage{booktabs}%
\usepackage{algorithm}%
\usepackage{algorithmicx}%
\usepackage{algpseudocode}%
\usepackage{listings}%
\usepackage{subcaption}  
\usepackage{float}
\usepackage{dsfont}

\theoremstyle{thmstyleone}%
\theoremstyle{thmstyletwo}%

\theoremstyle{thmstylethree}%
\newtheorem{definition}{Definition}%

\begin{document}

\title[Understanding possessions via path signatures]{The path to a goal: Understanding soccer possessions via path signatures}


\author*[1]{\fnm{David} \sur{Hirnschall}}\email{david.hirnschall@wu.ac.at}
\equalcont{These authors contributed equally to this work.}

\author[1]{\fnm{Robert} \sur{Bajons}}\email{robert.bajons@wu.ac.at}
\equalcont{These authors contributed equally to this work.}


\affil[1]{\orgdiv{Institute for Statistics and Mathematics}, \orgname{Vienna University of Economics and Business}, \orgaddress{\street{Welthandelsplatz 1}, \city{Vienna}, \postcode{1020}, \state{Vienna}, \country{Austria}}}

\abstract{We present a novel framework for predicting next actions in soccer possessions by leveraging path signatures to encode their complex spatio-temporal structure. Unlike existing approaches, we do not rely on fixed historical windows and handcrafted features, but rather encode the entire recent possession, thereby avoiding the inclusion of potentially irrelevant or misleading historical information. Path signatures naturally capture the order and interaction of events, providing a mathematically grounded feature encoding for variable-length time series of irregular sampling frequencies without the necessity for manual feature engineering. Our proposed approach outperforms a transformer-based benchmark across various loss metrics and considerably reduces computational cost. Building on these results, we introduce a new possession evaluation metric based on well-established frameworks in soccer analytics, incorporating both predicted action type probabilities and action location. Our metric shows greater reliability than existing metrics in domain-specific comparisons. Finally, we validate our approach through a detailed analysis of the 2017/18 Premier League season and discuss further applications and future extensions.}



\maketitle

\section{Introduction}\label{sec1}

A fundamental component of soccer analysis is the concept of possessions, which, broadly speaking, can be defined as consecutive sequences of actions executed by a single team. 
A game of soccer is naturally divided into these possessions. Hence, studying them has attracted a lot of interest recently. 
While several approaches have focused on evaluating players by estimating an outcome of possessions \citep{VanRoy20xT, Fernandez21EPV, Bajons25PMPoss}, a recently popularized approach is to evaluate possessions by predicting next actions. To this end, researchers have made a connection between natural language processing (NLP) and soccer events, which are both inherently sequential \citep{simpson2022seq2event,Mendes2024LEM,yeung2025transformer}. Hence, to solve the problem of predicting future actions, techniques predominantly used in NLP tasks such as recurrent neural networks (RNN, \citealp{Ellman90RNN}), gated recurrent units (GRU, \citealp{cho2014propertiesneuralmachinetranslation}), long-short-term memory (LSTM, \citealp{Hochreiter97LSTM}), and transformers (\citealp{vaswani2023attentionneed}). More specifically, \cite{simpson2022seq2event} use a transformer-based model architecture to predict next actions, whereas \cite{Zhang22LSTM_sports} use LSTMs to predict match results. Other approaches have made similar connections, but rely on either ordinary architectures such as sequentially concatenated multilayer perceptrons (MLP, \citealp{Mendes2024LEM}), or structures motivated by spatio-temporal point-processes \citep{yeung2025transformer}. While these approaches are intriguing, they ignore the nature of soccer possessions. Soccer possessions are spatio-temporal paths of different lengths, i.e., actually ordered time series where the length of the possession contains critical information. Hence, the previously mentioned approaches, which use a fixed historic window of past actions paired with sequential machine learning architectures, are an unnatural choice for this task. On the one hand, a randomly chosen fixed historic window of past actions may take into account actions from past possessions. That is, this approach almost inevitably takes into account multiple possessions, leading either to including both teams' actions or, if solely considering one team's action sequences, potentially substantial spatio-temporal discontinuities by omitting the opponents' intervening possessions. On the other hand, the fixed window of actions ignores the length of the current possession completely. A more natural way is to only include actions within one possession, or at least within a certain amount of past possession. However, this leads to the problem of time series of different lengths and irregular sampling frequencies, which the previously mentioned sequential architectures are not able to handle. In machine learning, this leads to the need for effective feature extraction techniques, aiming to identify salient characteristics of a time series informative for forecasting. Handcrafted features are usually based on domain-specific knowledge or general statistics, such as mean, variance, skewness, and kurtosis. While often useful, these features typically fail to take into account the temporal order of events, which is crucial in sports analytics.

To address these challenges, we propose to model possessions via path signatures. The signature of a path is a sequence of iterated integrals that gradually encodes spatio-temporal details until the path obtained by interpolation of data can, under mild conditions, be uniquely characterized. Informally, they can be viewed as feature extraction tool for time series that preserves the order and interaction of events over time. This makes signatures a natural choice for predicting actions. First, they are able to handle variable-length time-series, removing the need to fix a historic window size and avoid including potentially irrelevant or misleading information from distant possessions. Additionally, signatures retain all critical information without requiring geometrical or spatially handcrafted feature engineering, such as angles or differences used in previous works \citep{simpson2022seq2event,yeung2025transformer}. More precisely, our approach encodes all relevant information from the full possession by signatures to predict both the next action and its location in the form of $(x,y)$-coordinates. This approach allows us to evaluate possessions in detail, leading to various use cases for coaches. Furthermore, using signatures offers a general and mathematically valid alternative to existing approaches for a wide range of applications in sports with a similar spatio-temporal structure as soccer.

Building on our enhanced action prediction model, we devise a novel possession value metric, which takes into account both the predicted action type and the predicted location from our model. We provide evidence in favor of our possession value metric over existing ones and show that our proposed signature action prediction model is practically more relevant than the benchmark model from \cite{simpson2022seq2event}. We use our model and the possession evaluation metrics to get in-depth insight into the 2017/18 Premier League season. Furthermore, we complement this analysis with additional use cases for our model in practice. 

The main contributions of this paper can be summarized as follows:
\begin{itemize}
	\item We propose a novel and intuitive framework for action prediction in soccer using path signatures and only information from the recent possession.
    \item We show various advantages of our approach: (i) it is a natural approach for the complex spatio-temporal structure of possession, (ii) it allows the usage of time series of different lengths and sampling frequencies without the need for transformations to predict future actions, (iii) it is computationally more efficient, and improves prediction loss over existing approaches.
    \item We provide a novel way to evaluate possessions, taking into account action type probabilities and predicted locations in an intuitive and interpretable way. The metric is compared to existing ones in a domain-specific evaluation, outperforming them across various comparison setups.  
    \item We present a detailed analysis of the 2017/18 Premier League season and a number of practical applications for our action prediction model. 
\end{itemize}

\section{Literature review}

\subsection{Soccer modeling}
The advancements in data collection in recent years have revolutionized the way sports are analyzed. Nowadays, there is an increasing demand in professional sports for data scientists able to analyze highly granular data such as event stream and tracking data \citep{Lolli25survey}. Paired with advanced computational power and potent machine learning algorithms, the availability of new data has led to a vast range of data-driven analyses in the field of soccer analytics. 

While earlier work on analyzing soccer centered around modeling specific actions, such as passes \citep{McHale16passing,Horton17passing,Hvattum19passing}, or shots \citep{Robberechts20xG,Anzer21xG}, more recently, a focus has been on considering possessions, i.e., consecutive sequences of actions of one team, as unit of interest in soccer. \cite{VanRoy20xT} for example, compare and expected threat (xT) model to a model which values each action in a possession by estimating goal probabilities (VAEP). Both these models allow for explicitly assigning values to possessions and can be used for player evaluation. On the other hand, \cite{Fernandez21EPV} estimate expected possession value (EPV) to analyze various situations in soccer. The authors derive the EPV as a composition of different models, each of which is estimated via neural networks. \cite{Bajons25PMPoss} use possessions as units to estimate the player strengths via different regularized regression methods.

Due to the spatio-temporal nature of possessions in soccer, researchers found similarities between analyzing possessions and analyzing sentences in natural language processing (NLP). Hence, powerful methods with a proven success record in NLP, such as LSTMs and transformers, have been suggested for modeling sports data. In Basketball, \cite{Sicilia19LSTM_BB} use LSTMs to analyze possession, whereas \cite{Watson21rugby} use these models to analyze possessions in rugby union. In soccer, \cite{simpson2022seq2event} use transformers to account for the sequential nature of the game to analyze possessions. Their work is closest to our approach, and we use their model as a benchmark for our advanced action prediction model. Building on their work, \cite{yeung2025transformer} use a neural marked spatio-temporal point process (NMSTPP) architecture to model possessions. While they expand on \cite{simpson2022seq2event} by accounting for interevent time, they also consider a fixed historic window size and similar time series for the predictions of actions. Additionally, they refrain from directly modeling the $(x,y)$-coordinates of the next action and instead group the coordinates into predefined zones on the pitch. \cite{Mendes2024LEM} take a different route and derive a large event model (LEM), trying to predict a much more detailed set of 33 action types, location on a grid similar to \cite{yeung2025transformer}, and time elapsed on a prespecified grid. 

To the best of our knowledge, path signatures have not yet been used for modeling soccer possessions. Path signatures are, however, a natural choice for irregularly sampled spatio-temporal data such as possessions. Furthermore, they are able to handle possessions of different lengths while at the same time extracting all relevant information from possessions without the need for additional feature engineering. Due to their tremendous impact in other domains, as discussed in Section \ref{sec:sig_rev}, they have the potential to enhance models for various tasks in sports analytics. Our framework is only one example for the usage of signatures in sports, and the results of this work provide a glimpse of the power of signatures.

\subsection{Signatures for feature encoding}
\label{sec:sig_rev}
Signatures, originally introduced in the 50s \citep{chen1954iterated}, gained popularity in the mathematical community through their connections to the Rough Path theory developed by Terry Lyons \citep{lyons1998differential}. Recently, (log)-signatures have gained significant traction in machine learning applications as a non-parametric feature encoding and dimensionality reduction technique for time series data \citep{sturm2025pathsignaturesfeatureextraction}. Signatures are transformations that map a path to an infinite-dimensional sequence of statistics, capturing increasingly fine details of the underlying path. When appropriately augmented, the signature characterizes the path uniquely \citep{hambly2010uniqueness,boedihardjo2016signature}. They have been successfully utilized in various domains, including human action recognition tasks \citep{yang2017leveraging, yang2022developing} and gesture recognition \citep{gesture_sig}, predicting a diagnosis of Alzheimer’s disease \citep{moore2019using} and detecting early signs of depressive and manic episodes in patients with bipolar disorder \citep{kormilitzin2017detecting} as well as various applications in finance \citep{cuchiero2023signature, bayer2023optimal}.  
In the context of generative modeling, \cite{buehler2020generating} utilized path signatures paired with Variational Autoencoders to generate financial time series in a small data environment. \cite{liao2019learning} combined log-signatures with recurrent neural networks to learn neural stochastic differential equations and \cite{ni2021sig} measured time series similarities using a new metric based on signatures.

\section{Methodology}

\subsection{Path signatures}
When working with time series data, especially with irregularly sampled time intervals, as is typical for possession data streams, path signatures have emerged as powerful non-parametric feature maps. They provide a unique and concise representation of trajectories while encoding their structural properties in a mathematically principled way \citep{hambly2010uniqueness,chevyrev2016primer}. 

\begin{definition}{}
    For a continuous path with finite variation $X: [s,t] \rightarrow \mathbb{R}^d$ and a set of all multi-indexes $I=\left\{\left(i_1, \ldots, i_k\right) \mid k \geq 0, i_1, \ldots, i_k \in\{1, \ldots, d\}\right\}$ the signature is defined by a collection of iterated integrals of $X$, by
    \[S(X)_{s,t}=\left(1, S(X)_{s,t}^1, \ldots, S(X)_{s,t}^d, S(X)_{s,t}^{1,1}, S(X)_{s,t}^{1,2}, \ldots\right),\]
    with
    \[S(X)_{s, t}^{i_1, \ldots, i_k}=\int_{s<t_k<t} \ldots \int_{s<t_1<t_2} d X_{t_1}^{i_1} \ldots d X_{t_k}^{i_k}.\]
    By convention, the 0-th entry is equal to 1.
\end{definition}
The truncated signature of $X$ of order $M$ is defined as the finite collection of all terms $S(X)_{s,t}^{i_1, \ldots, i_M},$ where the maximum length of the multi-index is $M$. The truncation error at level $M$ decays with factorial speed as $\mathcal{O}(1/M!)$ \cite{lyons2007differential}. Computational examples illustrating how to compute the signature for simple paths are provided in \cite{chevyrev2016primer}.

To enrich the original possession stream, path augmentations are applied. In this work, we used a time augmentation and a visibility transformation, which add an extra dimension to encode information on the time stamps and the starting point, respectively. For more details on commonly used augmentations, we refer to \cite{morrill2020generalised}. Under appropriate augmentations, path signatures uniquely determine the path up to tree-like equivalences \citep{hambly2010uniqueness}.

Originally defined for continuous paths with bounded variation, the signature transformation has been extended to discrete paths by linear interpolation \cite{chevyrev2016primer}. For piecewise linear paths, computing signatures no longer involves integrals. Instead, according to Chen’s identity \citep{chen1958integration}, they can be calculated from the contributions of each line segment of the path, defined as
\[S(X)_{t, t+1}^{i_1, i_2, \ldots, i_k}=\frac{1}{k!} \prod_{j=1}^k\left(X_{t+1}^{i_j}-X_t^{i_j}\right),\]
where $X^{i_j}$ is the $i_j$-th coordinate of the path $X$.

Log-Signatures, denoted by $LogSig(X)_{s,t} = log(S(X)_{s,t})$ provide parsimonious representations of path signatures by significantly reducing their dimension.
Due to the shuffle product property of signatures \citep[Theorem 1.14]{chevyrev2016primer}, every polynomial function on signatures can be expressed as a linear combination of signature elements. This leads to repeated information in the signature representation, e.g. $S(X)_{s,t}^{i, i}=\frac{1}{2}\left(S(X)_{s,t}^i\right)^2.$ The log-signature essentially removes such redundancies, capturing the same information in fewer terms.

They are robust to irregularly sampled observations, where actions may not occur at uniform time intervals, unaffected by the length of the time series \citep{morrill2020generalised} and form a bijective map, uniquely determining the path up to tree-like equivalences \citep{hambly2010uniqueness}. For a detailed discussion of log signatures in machine learning, including their dimension reduction capabilities, we refer to \cite{liao2022log,morrill2020generalised}.

\subsection{Model architecture}
We process historical actions independently from other input features and propose log-signatures combined with weighted averages to effectively encode spatio-temporal structures of the underlying time series data. This enables the use of feed-forward neural networks instead of relying on computationally demanding transformer architectures, which yield faster training, improved scalability, and interpretability.

\begin{figure}[t!]
    \centering
    \includegraphics[width=\textwidth]{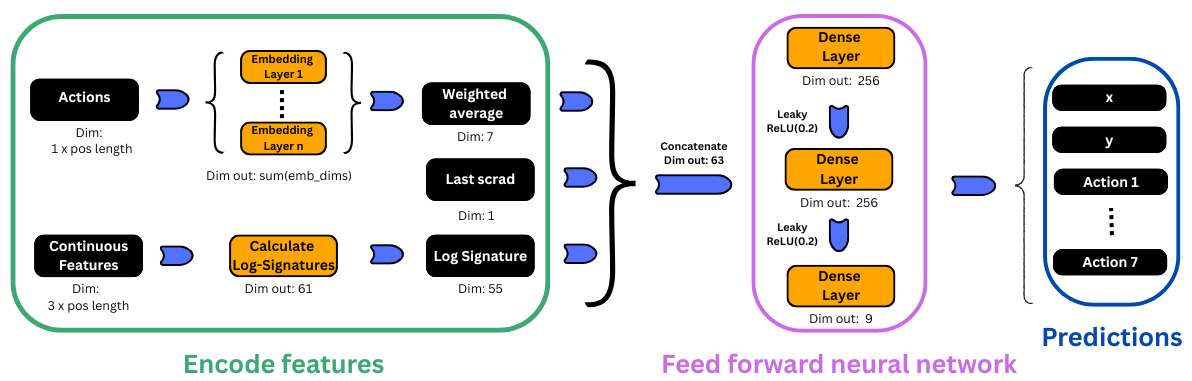}
    \caption{
    Sig-Model architecture in three stages. 
    Input actions are embedded and concatenated with the log-signatures of continuous features $(x,y,T)$ and the current scrad (score advantage) value. 
    Passing through the main neural network 
    yields predictions for both the next action location and the distribution of next action type.
    }
    \label{fig:model_plot}
\end{figure}

In contrast to previous work \citep{simpson2022seq2event,yeung2025transformer}, our model relies solely on previous actions, the latest difference in goals, and the raw triplets $(x,y,T)$, directly observed during a match. We deliberately avoid handcrafted geometric or temporal features, such as angles to the goal or action durations, as all information needed is naturally encoded by the path signature. Moreover, such features are often noisy, overly task-specific, or require extensive domain knowledge. Hence, if not carefully tailored to the model and task, they may introduce bias or overemphasis on specific aspects, potentially harming performance, as seen in Table \ref{tab:with_additional_features}, where we compare our model to a variant that includes the additional features mentioned in \cite{simpson2022seq2event}. 

More precisely, the time series of the continuous features $(x,y,T)$ are encoded as log-signatures, the sequence of actions is passed through an embedding layer, which maps each time step to a vector of dimension $emb\_dim$. To account for the temporal structure and the decreasing importance of features over time, we compute a weighted average, assigning higher weights to more recent actions. The weights are set to the inverse of each action's position in the sequence and are scaled to sum to one, thereby emphasizing more recent events while still incorporating information from earlier actions. 

Both components are then concatenated together with the current score advantage, i.e., number of goals ahead or behind (referred to as $scrad$), and passed through a feed-forward neural network, as illustrated in Figure \ref{fig:model_plot}.

The network contains one input, one output, and one hidden layer of 256 nodes and utilizes leaky rectified linear unit (Leaky ReLU) activations with parameter $\alpha=0.2$ between layers. The output vector of length 9 is split into action logits of length 7 and $x$ and $y$ predictions. Our proposed model pipeline in its used hyperparameter configuration is illustrated in Figure \ref{fig:model_plot}. Details on the performed hyperparameter grid search can be found in \ref{sec:tuning}.

Following \cite{simpson2022seq2event}, we use a two-part cost function.
The error of the predicted ball's location is measured by the root mean squared error (RMSE) in $x$ and $y$. Predicted action probabilities are evaluated using a weighted cross-entropy loss (CEL), where goals, change of possession events, and end-of-match events are considered as purely contextual rather than indicators of style, hence are given zero weights within the CEL function. As final cost function, we use the weighted sum 
\begin{align}\label{cost_function}
    L(\theta) = RMSE_{(x,y)} +\lambda CEL_{actions}, 
\end{align}
where $\lambda$ is treated as a hyperparameter.

\subsection{Empirical evaluation}

In this section, we study the performance of our proposed methodology in various settings and compare it to a benchmark model from the literature. All computations were performed on a 2023 MacBook Pro equipped with an M3 Pro chip, 36GB of RAM, and 12 CPU cores. Code for reproducing all results is available at \url{https://github.com/Rob2208/sig_actions}.

\subsubsection{Data and preprocessing}
\label{sec:data_pp}
For model evaluation, we utilize WyScout Open Access Dataset \cite{pappalardopublic}, including match data from the 2017/18 season of the top men's leagues in England, Germany, France, Italy, and Spain, enhanced by matches from the Euro 2016 and World Cup 2018. For the comparison of models in Section \ref{sec:bench}, we follow \cite{simpson2022seq2event} and select 138 out of the total 1941 matches across leagues and success rates. However, we perform various permutations of this approach in additional validations of the model in Appendix \ref{appendix_hyperparameter}.

Opposed to existing work, we are not stipulating an arbitrary historic window of actions, but rather follow a more natural approach and treat every possession as its own self-contained sequence following tactical considerations, to align with how coaches and analysts evaluate the game. This approach is enabled by using path signatures, as they encode spatio-temporal properties regardless of a sequence's length and sampling frequency. Further, we will only rely on input features that are directly observable during a match: performed actions, current goal difference, and the raw location-time triplet $(x,y,T)$ scaled by their respective maximum value. Hence, we deliberately avoid hand-crafted geometric and temporal features such as shot angles, action durations, or spatial differences in $x$ and $y$ coordinates.
To support our choice of pre-processing and for comparison purposes, we examined how the length of the historic window and the exclusion of additional features impact the Seq2Event transformer model from \cite{simpson2022seq2event}. As shown in Table \ref{tab:transformer_losses}, the length of the historic window appears to be arbitrarily long without a clear indication of the optimal value. Furthermore, removing all handcrafted features and relying solely on observable features $(x,y,T)$, yields a clear reduced prediction accuracy of the Seq2Event transformer model, indicating a need for additional geometrical features when using their approach (see again Table \ref{tab:transformer_losses}). 

We evaluate our model's forecasting ability by starting at different points within a possession, namely 4th, 5th, 6th, 7th, or 8th action. Beginning to forecast only when at least a certain number of actions have occurred not only ensures that enough information for a realistic forecast is provided but also filters out short possessions, which, in general, provide little tactical context, hence are not meaningful for evaluating possession utilities. 

For each test possession, we form a path $(x,y,T)$, apply a time and an invisibility-reset augmentation, followed by a piecewise linear interpolation, and calculate the log-signature of order 3. The order is chosen based on the hyperparameter tuning results in \ref{sec:tuning}. Categorical action features are embedded directly in our proposed model. 
In order to ensure comparability to previous works \citep{simpson2022seq2event, yeung2025transformer}, actions were grouped into seven categories: pass (‘p’), dribble (‘d’), cross (‘x’), shot (‘s’) and goal scored (‘g’), possession end (‘$\_$’), and match end (‘@’). The ‘@’ category is only a helper category in order to distinguish between matches, but is not relevant for prediction. While there are different approaches to group actions (see e.g. \citealp{Mendes2024LEM}), we agree with previous work that this is a suitable representation of relevant soccer actions.  

\subsubsection{Evaluation metrics}
Model evaluation is conducted on four losses. We present the test loss \eqref{cost_function} split in its components, the RMSE in $x$ and $y$, and a weighted CEL, to evaluate the predicted location and action probabilities, respectively.

To assess the full predictive distribution of actions $A$ with values $a \in \mathcal{A}$, we investigate the Brier score 
\begin{equation}
\label{eq:BS}
    \text{Brier Score} = \frac{1}{N} \sum_{i=1}^{N} \sum_{k=1}^{K} (f_{i,k} - \mathds{1}_{\{a_i=k\}})^2,
\end{equation}
where $f_{i,k}$ denoted the predicted probability that action $a_i$
 belongs to class $k$.
Moreover, the Kullback-Leiber (KL) divergence between the predicted and the empirical zone-conditioned action distribution is presented. For actions $A$ 
occurring in zone $z_i$ following the empirical distribution $\mathbb{Q}_{z_i}=(q_{z_i}(a))_{a\in\mathcal{A}}$ and predicted distribution $\mathbb{P}_{z_i}=(p_{z_i}(a))_{a\in\mathcal{A}}$ the KL divergence is given by \\
\begin{equation}
\label{eq:KL}
    KL(\mathbb{P},\mathbb{Q}) = \sum_{a\in\mathcal{A}} p_{z_i}(a)log\frac{p_{z_i}(a)}{q_{z_i}(a)}.
\end{equation}

\begin{figure}[t!]
    \centering
    \includegraphics[width=\textwidth]{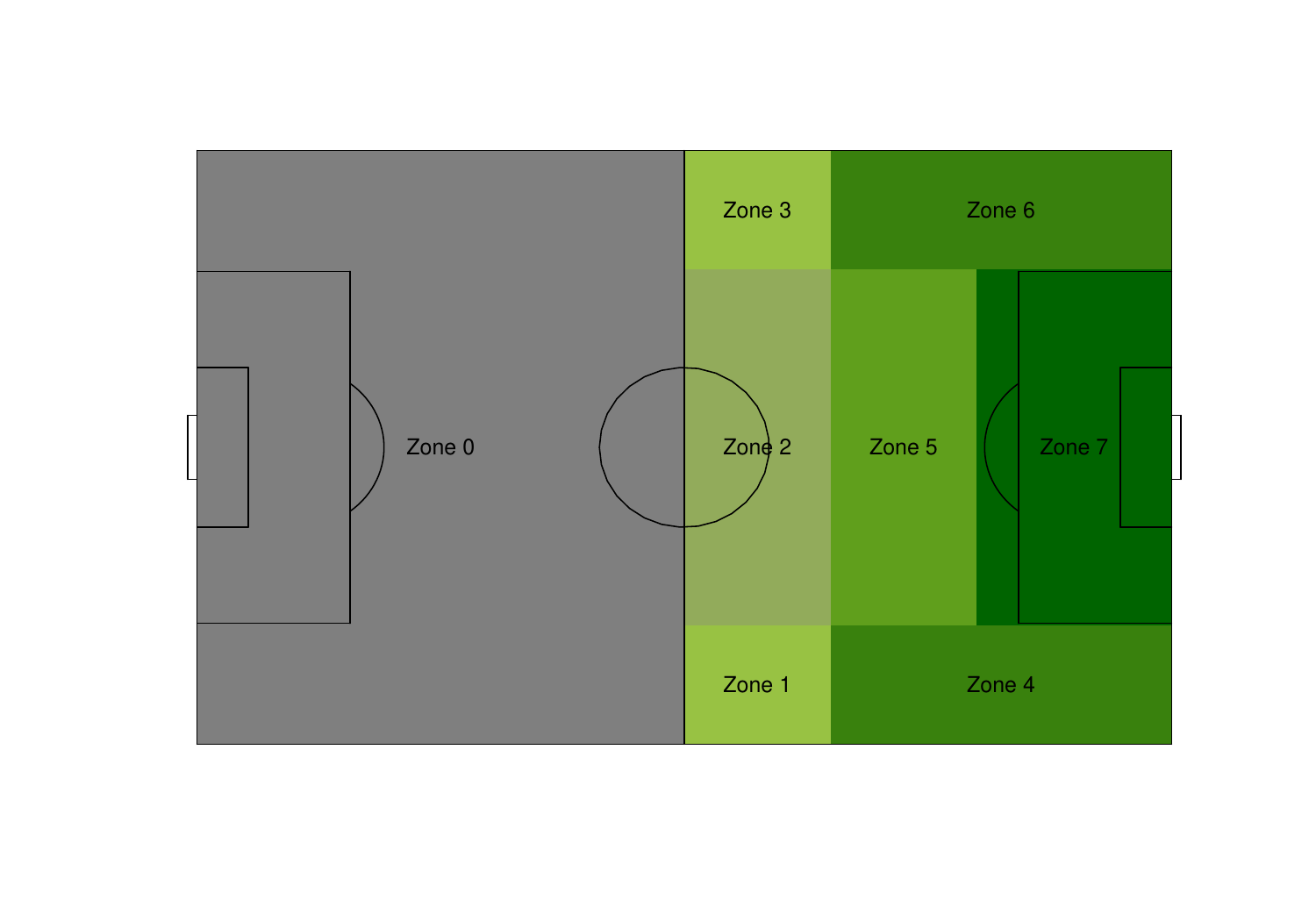}
    \caption{
    A standard soccer pitch divided into 8 relevant zones. For teams in possession, the play direction is from left to right. 
    }
    \label{fig:zones}
\end{figure}

Whereas CEL and Brier score reward selecting the correct action, we include the KL divergence to measure the tactical plausibility of predicted actions. It shows whether a model is in line with typical soccer-specific expectations in the given pitch zone. 
The zones are chosen according to Figure \ref{fig:zones}. We used domain knowledge to create the zones and focused on the offensive side of the pitch, as more relevant actions happen in this half. The zones are chosen such that they clearly favor a specific set of actions. Looking at Figure \ref{fig:zones}, we expect more shots in Zone 7 than in other zones, whereas we expect more crosses in zones 4 and 6. This information is also contained in the empirical zone distribution. Hence, a model that is close to this distribution (in terms of KL divergence) and at the same time performs well in the other evaluation metrics indicates a good balance between predictive accuracy and tactical reasonability. 

\subsubsection{Hyperparameter search}\label{sec:tuning}
A comprehensive model selection and hyperparameter grid search was performed, evaluating every parameter combination on four loss metrics across various datasets. Models were scored on possession sequences using $n_r\in\{3, 4, 5, 6, 7\}$ recent action features, which corresponds to forecasting from the 4th, 5th, 6th, 7th or 8th action onwards. In total, the search procedure resulted in 480 different configurations across all combinations of hyperparameters, data splits and values of $n_r$.
We conduct the hyperparameter tuning process in two stages, with tested parameters listed in Table \ref{tab:params}. In the first phase, we optimize the CEL scaling factor $\lambda$, the signature order $M$, and the batch size $n_{batch}$ (in that order). 
Each parameter was evaluated on the initial dataset described in \ref{sec:data_pp}. The best value for each hyperparameter was selected in a robust way based on the average losses across the grid of the remaining hyperparameters. 
While Table \ref{tab:cel}, shows a superior performance for $\lambda=1$ across all $n_r$, Table \ref{tab:sig} indicates the best performance for signature order $M=3$. Although higher signature order lowered the KL for longer possessions, $M=3$ achieved the best overall trade-off and was chosen. Finally, a batch size of $n_{batch}=32$ slightly decreased the Brier score but significantly increased the RMSE, hence it was discarded (see Table \ref{tab:batch}). 

In the second phase, we retained the six most promising hyperparameter combinations from stage one 
as model candidates (see Table \ref{tab:model_configs}) and assessed their robustness. For this, we train and test each model across 10 randomly generated train-test splits, obtained by varying sets of teams and matches included in the corresponding datasets. 

Average scores across the generated data splits for each model candidate are presented in Table \ref{tab:mean_std_all_models} separately for each value of $n_r$. The final loss comparison resulted in a signature order of $M=3$, a CEL scaling factor of $\lambda=1$, a batch size of $n_{batch}=4$, and a hidden dimension of 256.

\subsubsection{Comparison with baseline model}\label{sec:bench}
Finally, we benchmark our proposed model (Sig-Model) against the transformer-based Seq2Event model in \cite{simpson2022seq2event}. For comparison reasons, we report losses and runtimes in seconds for the identical set of 138 matches as used by \cite{simpson2022seq2event} (see \ref{sec:data_pp}). 
Results are presented in Tables \ref{tab:loss_comparison} and \ref{tab:runtimes} respectively.

\begin{table}[h]
    \centering
    \caption{Losses of our Sig-Model and the Seq2Event model for $n_r\in\{3,4,5,6,7\}$, where Test loss = MSE + CEL on the test dataset. Best values are highlighted in bold.}\label{tab:loss_comparison}
    \begin{tabular}{c c r r r r r}
        \toprule
        \textbf{$n_r$} & \textbf{Model} & \textbf{Test loss} & \textbf{MSE} & \textbf{CEL} & \textbf{Brier} & \textbf{KL} \\
        \midrule
        \multirow{2}{*}{3}
            & Sig-Model   & \textbf{0.2084} & \textbf{0.1598} & 0.0486 & \textbf{0.7968} & \textbf{0.1117} \\
            & Seq2Event   & 0.2129          & 0.1652          & \textbf{0.0477} & 0.8034 & 0.1121 \\
        \midrule
        \multirow{2}{*}{4}
            & Sig-Model   & \textbf{0.2096} & \textbf{0.1601} & 0.0495 & \textbf{0.7938} & \textbf{0.1125} \\
            & Seq2Event   & 0.2144          & 0.1658          & \textbf{0.0486} & 0.8054 & 0.1163 \\
        \midrule
        \multirow{2}{*}{5}
            & Sig-Model   & \textbf{0.2108} & \textbf{0.1601} & 0.0507 & \textbf{0.7877} & \textbf{0.1146} \\
            & Seq2Event   & 0.2142          & 0.1652          & \textbf{0.0490} & 0.7984 & 0.1177 \\
        \midrule
        \multirow{2}{*}{6}
            & Sig-Model   & \textbf{0.2121} & \textbf{0.1606} & 0.0515 & \textbf{0.7858} & \textbf{0.1155} \\
            & Seq2Event   & 0.2141          & 0.1646          & \textbf{0.0495} & 0.7954 & 0.1183 \\
        \midrule
        \multirow{2}{*}{7}
            & Sig-Model   & \textbf{0.2134} & \textbf{0.1612} & 0.0522 & \textbf{0.7881} & \textbf{0.1169} \\
            & Seq2Event   & 0.2152          & 0.1642          & \textbf{0.0510} & 0.7998 & 0.1202 \\
        \bottomrule
    \end{tabular}
\end{table}
Our model outperforms the benchmark in the test loss obtained by the cost function \ref{cost_function}, as it reduced the MSE significantly while almost maintaining the CEL. Further, the Sig-model achieved a higher Brier score and a lower deviation from the empirical zone-conditioned action distribution in terms of KL divergence across all forecasting starting points in the given possessions.
\begin{table}[h]
    \centering
    \caption{Runtimes in seconds of our Sig-Model and the Seq2Event model for each $n_r\in\{3,4,5,6,7\}$. Best values are highlighted in bold.}\label{tab:runtimes}
    \begin{tabular}{c c r  }
        \toprule
        \textbf{$n_r$} & \textbf{Model} & \textbf{Runtime in sec}  \\
        \midrule
        \multirow{2}{*}{3}
            & Sig-Model & \textbf{280.67} \\
            & Seq2Event & 687.97 \\
        \midrule
        \multirow{2}{*}{4}
            & Sig-Model & \textbf{253.30} \\
            & Seq2Event & 391.62 \\
        \midrule
        \multirow{2}{*}{5}
            & Sig-Model & \textbf{237.21} \\
            & Seq2Event & 383.09 \\
        \midrule
        \multirow{2}{*}{6}
            & Sig-Model & \textbf{227.45} \\
            & Seq2Event & 323.51 \\
        \midrule
        \multirow{2}{*}{7}
            & Sig-Model & \textbf{194.71}\\
            & Seq2Event & 250.31 \\
        \bottomrule
    \end{tabular}
\end{table}

On the tested dataset, the runtimes were significantly reduced. Especially when starting forecasting early in the possession, i.e., with an increased number of forecasts, we observe a decrease in runtime of a factor of about 2.5. Note that the previously reported runtime of approximately 45 minutes for the Seq2Event model \citep{simpson2022seq2event,yeung2025transformer} was reduced to the values reported in Table \ref{tab:runtimes}, due to technical improvements in the Pytorch implementation, particularly in how training samples are stored and loaded.

\section{Practical application}
\label{sec:practice}

\subsection{Possession value}
A popular use case for action prediction models is to evaluate possessions. Due to the dynamic nature of soccer, it is difficult to assess the value of a possession sequence as, in general, most possessions lack a valuable outcome, such as a goal. \cite{simpson2022seq2event} assess possessions by analyzing the attacking weight of a possession. Intuitively, a possession that ends in, or contains, an attacking action (regardless of whether being successful or not) is more valuable than a possession without attacking intent. Hence, \cite{simpson2022seq2event} assign a value to possession based on the cumulative predicted probabilities for the actions shot and cross, which can be seen as an attacking event. Their poss-util metric for a possession $p$ is defined as
\begin{equation}\label{eq:poss-util} 
\operatorname{poss-util} = \sum_{i=1}^{N_p} P_i(\text{cross, shot})\cdot c, 
\end{equation} 
where $N_p$ is the number of actions in possession $p$, $P_i(\text{cross, shot})$ is the predicted probability of a shot or cross for action $i$, and $c \in \{-1,1\}$ is a factor depending on whether a shot, or cross actually happened or not. While poss-util allows for insights into a team's effectiveness and playing style, it is not a natural choice of metric for an action prediction model that outputs predictions for actions and locations on the pitch. Leveraging the information on the location of the actions within a possession allows for a more detailed evaluation of possession. To this end, \cite{yeung2025transformer} develop the HPUS metric for possessions:
\begin{equation}\label{eq:HPUS} 
\operatorname{HPUS} = \sum_{i=1}^{N_p} \phi(N_p+1-i) \operatorname{HAS}_i. 
\end{equation} 
HPUS is a weighted average of action values (HAS) for each action, where the weighting function $\phi$ emphasizes actions that happen later during the possession. The action score depends on a value for the predicted action type, a value for the predicted location, and the elapsed time between actions. In more detail, HAS is defined as
\begin{equation}\label{eq:HAS} 
\operatorname{HAS} = \frac{\sqrt{\operatorname{AV}\cdot \operatorname{ZV}}}{t}, 
\end{equation} 
where the action value (AV) is given as 
\begin{equation}\label{eq:AV} 
\operatorname{AV} = 0\cdot P(\text{possession loss}) + 5\cdot P(\text{dribble, pass}) + 10\cdot P(\text{cross, shot}),
\end{equation} 
and similarly, the location value (ZV) depends on zones and is given as
\begin{equation}\label{eq:ZV} 
\operatorname{ZV} = 0\cdot P(\text{Area}_0) + 5\cdot P(\text{Area}_1) + 10\cdot P(\text{Area}_2),
\end{equation} 
and $t$ represents the interevent time. While HPUS makes use of the location and the elapsed time of the actions (additional to action types), it is not clear how the factors (0, 5, and 10) and the zones ($\text{Area}_0$, $\text{Area}_1$, and $\text{Area}_2$) were chosen. In fact, \cite{yeung2025transformer} even suggest that these values can be adjusted arbitrarily. However, this makes HPUS difficult to interpret and also questions the reliability of the metric, as different values could potentially lead to different results.

To address the above deficiencies, we devise a new metric for possession value. Our metric is built upon well-known and widely used metrics for evaluating actions. We follow a similar approach to HPUS and assign values to each action in a possession. However, instead of multiplying values for the location on top of that, we make the value location-dependent. In particular, our location-based possession value (LPV) assigns a location-based action value (LAV) to each action of a possession as
\begin{equation}\label{eq:act_val} 
\operatorname{LAV} = \widehat{\operatorname{xG}} \cdot P(\text{shot}) + \widehat{\operatorname{xT}} \cdot P(\text{dribble, pass, cross)},
\end{equation} 
where $\widehat{\operatorname{xG}}$ is a simple expected goals model \citep{Robberechts20xG,Anzer21xG} evaluated at the predicted $(x,y)$-location, and $\widehat{\operatorname{xT}}$ is an expected threat model \citep{Singh19xT,VanRoy20xT} again evaluated at the predicted $(x,y)$-location. We provide more details on the computation of these models in Appendix \ref{app:pvd}. The proposed DPV for the full possession is then computed similarly to before by simply summing over all actions in a possession. 
\begin{equation}\label{eq:poss_val} 
\operatorname{LPV} = \sum_{i=1}^{N_p} \operatorname{LAV}_i.
\end{equation} 
Using \eqref{eq:poss_val} to obtain a value for a possession thus takes into account the action type predictions, while at the same time using the predicted location to assign each action a value that is interpretable in terms of widely used metrics in soccer. Finally, we briefly discuss some possible adaptations of \eqref{eq:act_val} and \eqref{eq:poss_val}. First, in comparison to HPUS, we do not take the elapsed time between actions into account. While the intuition from \cite{yeung2025transformer} that faster actions may lead to more threatening possessions seems reasonable, they do not provide a data-driven justification for that claim. To maintain comparability in model comparison (compare section \ref{sec:bench}), we did not explicitly model the time an action takes. While it would be relatively straightforward to extend our framework to include time, we leave it open for future work. Second, we also did not incorporate a weighting function $\phi$ as in \eqref{eq:HPUS}. Again, we believe that the idea from \cite{yeung2025transformer} is sensible, but it is not clear whether their choice of $\phi$ is suitable. We believe that a weighting function is more important for lengthier possessions, whereas for short possessions, all actions may be equally valuable. Hence, an adequate function $\phi$ should be carefully derived. Since this is not the purpose of this work, we leave this for future work as well.  

\subsubsection{Choosing a possession value metric}
\label{sec:choosing_pv}

In this section, we attempt to validate our choice of possession value metric. First, in order to analyze which possession value may be the most suitable, we have to establish how metrics of the form \eqref{eq:poss-util}, \eqref{eq:HPUS}, and \eqref{eq:poss_val} can be used in practice. In their respective articles, \cite{simpson2022seq2event} and \cite{yeung2025transformer} use their prediction models to obtain values for each possession. While this approach provides insights into a team's performance and potentially its playing style, it omits an important aspect, namely, what actually happened. In our opinion, the true strength of using action prediction models for deriving possession values relies on comparing the models' predictions with the actually observed possessions. Indeed, for all three metrics presented above, an ``observed'' possession value can be computed straightforwardly. Instead of plugging in the models' probabilities for each action type into \eqref{eq:poss-util}, \eqref{eq:HPUS}, and \eqref{eq:poss_val}, we simply insert a 1 for the action type that actually took place and a 0 for the other types. To be more precise, in the case of \eqref{eq:poss-util} (poss-util), we hence count the number of crosses and shots in a possession. In the case of \eqref{eq:HPUS} (HPUS), \eqref{eq:AV} (AV) is either 0, if the actual action was a ``possession loss'', 5 if the actual action was a ``dribble'' or ``pass'', and 10 if it was a ``cross'' or ``shot''. Similarly, \eqref{eq:ZV} (ZV) is 0, if the actual $(x,y)$-coordinates correspond to a location in $\text{Area}_0$, 5 if they correspond to a location in $\text{Area}_1$, and 10 if the actual action happened in $\text{Area}_2$\footnote{Note that, for simplicity, we ignore time and set $t = 1$ in \eqref{eq:HAS} for all actions.}. For our LPV metric, \eqref{eq:act_val} is equal to an xG value at the observed location of the action if it was a shot, and equal to an xT value at the observed location if it was any other action (different from a ``possession loss''). 

\begin{figure}[t!]
    \centering
    \includegraphics[width=\textwidth]{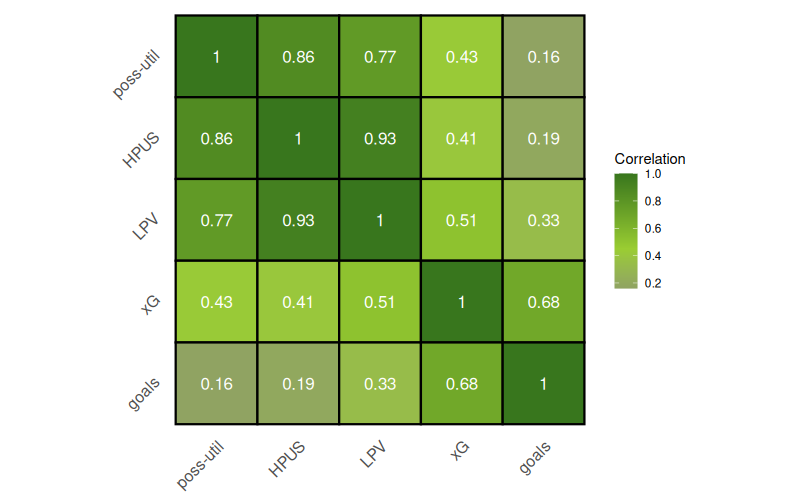}
    \caption{
    Correlation matrix of different possession value metrics (poss-util, HPUS, and LPV) and relevant outcome indicators (xG and goals) for each match of the 2017/18 Premier League season.
    }
    \label{fig:corrs1}
\end{figure}

One way to assess which metric is most suitable for capturing possession value is by examining its correlation with actual game outcomes. In particular, we analyzed all games of the 2017/18 season of the Premier League. For each game, we computed the total possession value of each team, i.e., we aggregated the possession values of each possession they had, and correlated them with goals and xG values in that same match. To be precise, we obtained goals and xG values, i.e., the aggregated xG values for each shot, for each match from \href{https://understat.com/}{understat}, a company providing xG values and results for various leagues. These values are obtained using the \texttt{worldfootballR} package in \texttt{R}. The main idea is that a useful possession value metric, while certainly providing distinct information, should be at least to some degree correlated to relevant outcome indicators \cite{Davis24metric_eval}.  
Figure \ref{fig:corrs1} shows the results of this analysis. On the one hand, we note that all 3 possession value metrics are highly correlated. Furthermore, LPV is more strongly correlated with HPUS than poss-util. This is expected as both take into account location information. On the other hand, we observe that LPV has a higher correlation with xG and goals than the other metrics, suggesting a higher correlation with relevant outcome indicators compared to the alternatives. Hence, our LPV metric is able to capture context relevant for successful teams. Note that the correlation between our metric and goals is nevertheless lower than the correlation between xG and goals. However, the aim of our LPV metric is to measure possession value, and a team can have high possession value, but score few goals. Lastly, it should be noted that comparison of our LPV with an xG may seem unfair, as it also contains an ``xG'' term. However, the xG term in LPV is based on a very simple model (see Appendix \ref{app:pvd}), whereas modern xG models take into account a wide range of features \citep{Robberechts20xG,Anzer21xG}. Although it is not entirely clear how the xG value from understat is computed, it is mentioned that they model xG values via a neural network trained on a large dataset. This is in contrast to the simple xG methodology used for LPV as described in Appendix \ref{app:pvd}. Hence, we believe it is sensible to include the comparison with the external xG from understat in addition to comparisons with actual goals. 

\begin{figure}[t!]
    \centering
    \includegraphics[width=\textwidth]{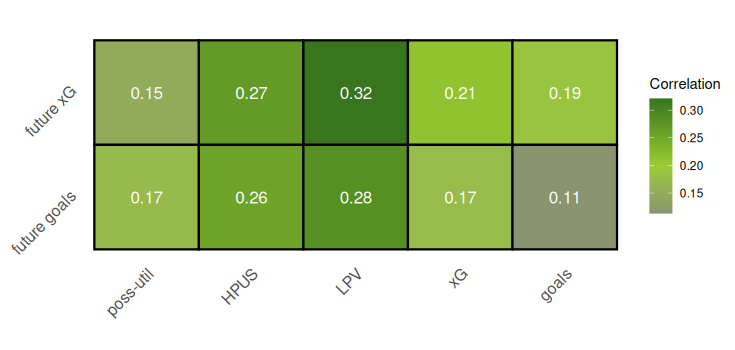}
    \caption{
    Comparison of the possession value models (poss-util, HPUS, and LPV) and relevant outcome indicators (xG and goals) with future performance. Displayed is the correlations between poss-util, HPUS, LPV, xG, and goals from one match of a team and xG and goals in the subsequent game of the team (denoted by future xG and future goals). 
    }
    \label{fig:corrs2}
\end{figure}

Another popular approach for assessing metrics is to correlate them to future results \citep{HvattumGelade21,Davis24metric_eval}. In particular, we follow a similar approach as \cite{Spearman18beyond_xG}, and correlate the possession value metrics to future performance in terms of goals and xG, to check whether our metric is more predictive than scoring itself. To do so, we compute the correlation between possession value metrics, goals, and xG to goals and xG in the subsequent match on a team basis. Figure \ref{fig:corrs2} displays the results of this analysis. In general, HPUS and our LPV show higher correlation with future performance than xG and goals themselves. In addition, our LPV is the most correlated with future performance among all other metrics. 

In conclusion, the results presented in Figures \ref{fig:corrs1} and \ref{fig:corrs2} strengthen our belief that the possession value metric devised in this paper is superior to HPUS and poss-util. 

\subsubsection{Comparison of prediction models in practice}

After establishing a practicable possession value metric, we can turn to analyzing prediction models from a domain-specific viewpoint. In Section \ref{sec:bench}, we compared the performance of our Sig-Model to a benchmark action prediction model from \cite{simpson2022seq2event} in terms of runtime and loss on a test set. In this section, we again compare our Sig-Model to their Seq2Event model. In particular, we aggregate the predicted possession values over each match and compare them to future performances, as obtained by the aggregated LPV in the subsequent game. Similarly to before, a higher correlation between values from one match and the next match is a desirable characteristic of a performance metric \cite{Davis24metric_eval}. 

\begin{figure}[t!]
    \centering
    \includegraphics[width=\textwidth]{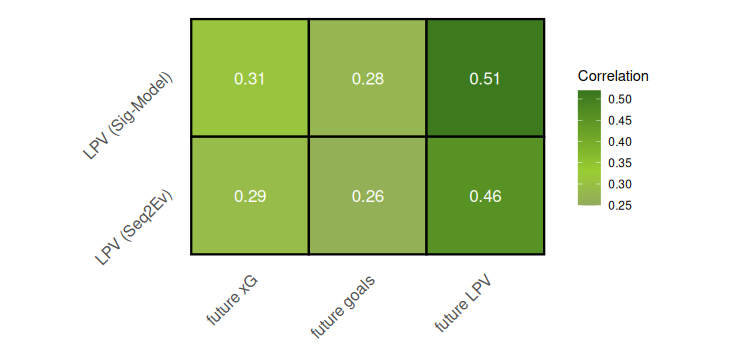}
    \caption{
    Comparison of Sig-Model and Seq2Event model. For each match from a team, the (aggregated) LPV is calculated and the correlation with the xG, goals and (aggregated) LPV in the subsequent match of the team is computed.
    }
    \label{fig:corrs3}
\end{figure}

Figure \ref{fig:corrs3} shows the correlation of LPV with future xG, future goals, and with a future LPV. The values for LPV are computed once with our Sig-Model and once with the transformer-based Seq2Event model from \cite{simpson2022seq2event}. In all three cases, we observe higher correlation with future performance for the Sig-Model than for the Seq2Event model. This result reinforces the findings from Section \ref{sec:bench} and provides confidence for the use of our framework in practice. 

\subsection{Application to Premier League}
\label{sec:PL}

\begin{figure}[t!]
    \centering
    \includegraphics[width=\textwidth]{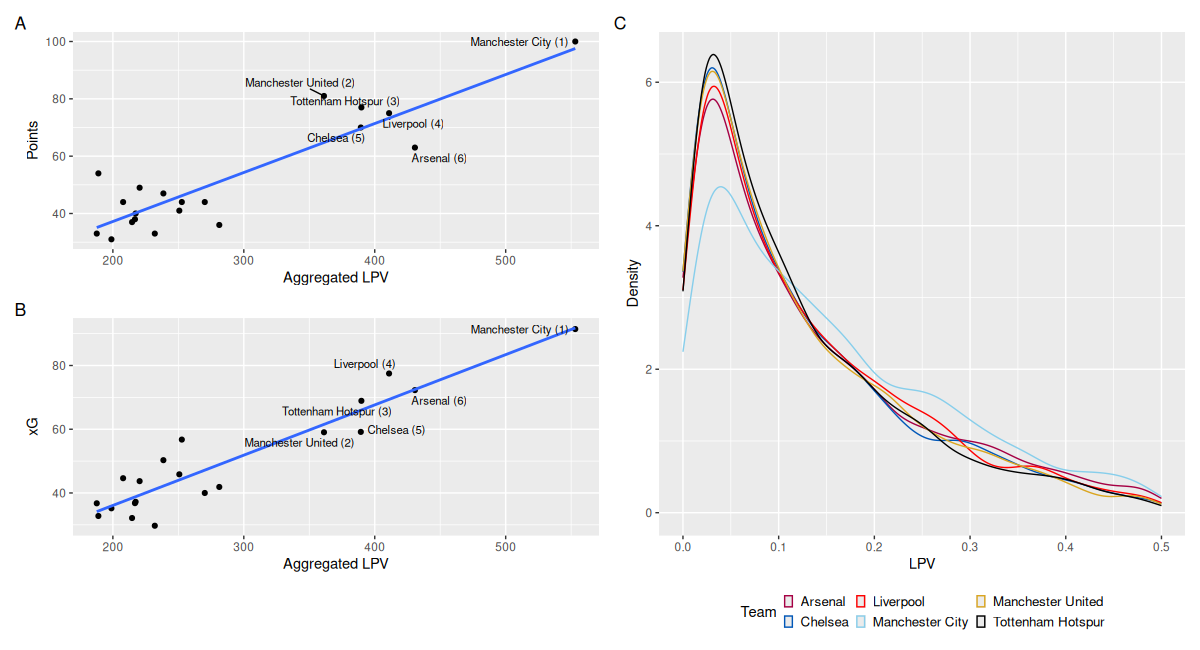}
    \caption{
    Analysis of the LPV values for the 2017/18 Premier League season. Panel A displays a scatterplot of the aggregated LPV value of a team over the season and the total final points. Panel B shows a scatterplot of the aggregated LPV value of a team over the season and the aggregated xG values of a team over the season. Panel C shows density plots of LPV for the top 6 teams in that season.  
    }
    \label{fig:use_case1}
\end{figure}

To highlight the potential of our action prediction model and the LPV metric, we analyze the 2017/18 Premier League season through the lens of the Sig-Model. To this end, we first compare teams by means of LPV. Figure \ref{fig:use_case1} shows three different plots connecting a team's points, xG, and possession value. Panel A shows the relationship between a team's final points and the accumulated LPV. We observe a strong positive correlation ($R \approx 0.896$), indicating that teams with high possession value performed better in terms of final ranking. Indeed, the top-performing team of the 2017/18 Premier League season, Manchester City, also had the highest cumulative possession value over the season. Notably, Panel A reveals three distinct clusters of teams. First, Manchester City performed far better than the next competitors (in terms of points and possession value). Then, there is a visible cluster of teams ranked in places 2-6 in the final league table (Manchester United, Rank 2; Tottenham Hotspur, Rank 3; Liverpool, Rank 4; Chelsea, Rank 5; and Arsenal, Rank 6). Finally, there is a cluster for the rest of the teams. Among weaker teams, the correlation between cumulative possession value and total points is lower. Panel B displays the correlation between accumulated xG values and accumulated possession values. There is an even stronger positive correlation observable in Panel B ($R \approx 0.934$). In fact, we observe that the alignment between accumulated xG values and accumulated possession values is more pronounced also for weaker teams. From a domain-specific viewpoint, this result is expected, as the possession value metric and xG both are measures of offensive quality, whereas final points to a certain degree account for offensive and defensive team strength. A particular example is Manchester United, the runner-up of that season. They played an outstanding season from a defensive viewpoint, conceding only 28 goals, the second fewest this season after leader Manchester City (27 goals conceded). Offensively, however, they performed worse, ranking only 5th and 6th in total goals scored and xG created, respectively. This can be attributed to the playing style of José Mourinho, Manchester United's coach in that season, who is well-known for playing pragmatic and result-oriented football based on a solid defensive line\footnote{See for example this \href{https://www.sofascore.com/news/famous-football-managers-and-their-philosophies-jose-mourinho/}{article}, or this \href{https://www.youtube.com/watch?v=1JBDCilBnas&t=18s}{Youtube video}.}. On the other hand, Arsenal, ranked only 6th in the 2017/18 final league table, had an offensively strong season but struggled defensively, another well-studied fact\footnote{See for example this \href{https://arseblog.news/2018/05/arsenal-2017-18-by-the-numbers/}{article} reviewing Arsenal's season by common soccer stats.}. Hence, our metric is able to capture playing style of different teams adequately, at least in terms of offensive style. Panel C displays the densities for the possession values of the top 6 teams. The story is similar to before: Manchester City is much more threatening, showing a higher density on high possession values; Arsenal (6) and Liverpool (4), both offensively strong teams follow with higher right tails; Manchester United (2) and Chelsea (5), defensively strong teams, have more mass at smaller possession values.

\begin{figure}[t!]
    \centering
    \includegraphics[width=\textwidth]{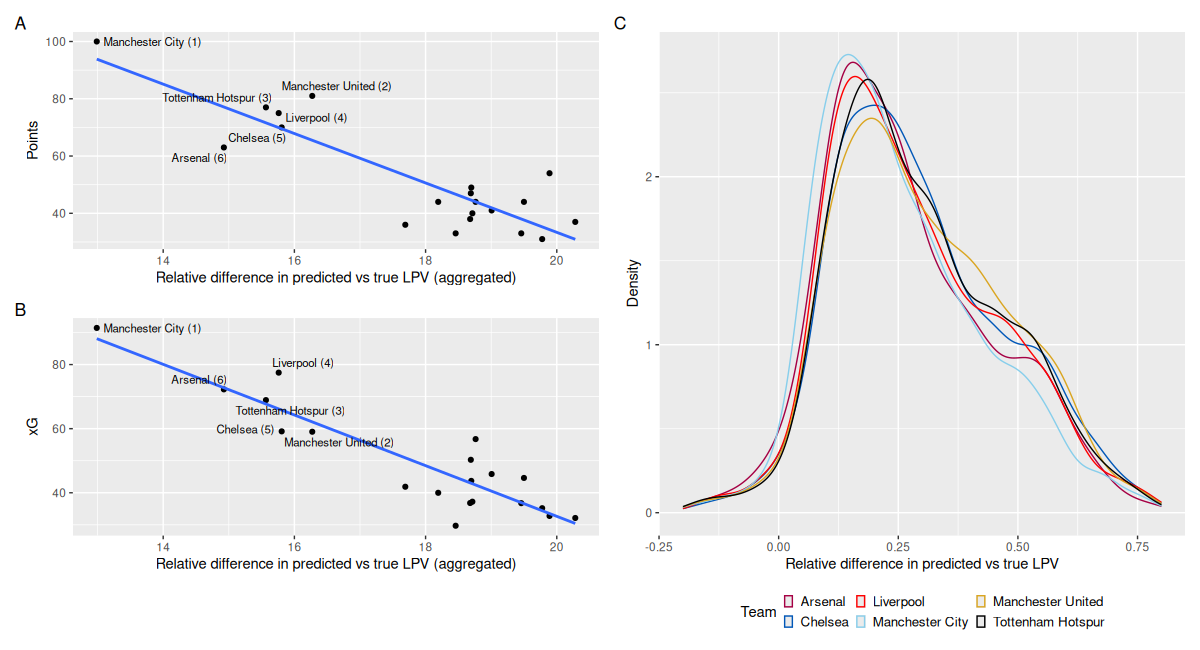}
    \caption{
    Analysis of the relative difference in predicted vs true LPV values for the 2017/18 Premier League season. Panel A displays a scatterplot of the aggregated relative difference in predicted vs true LPV value of a team over the season and the total final points. Panel B shows a scatterplot of the aggregated relative difference in predicted vs true LPV value of a team over the season and the aggregated xG values of a team over the season. Panel C shows density plots of the relative difference in predicted vs true LPV for the top 6 teams in that season.  
    }
    \label{fig:use_case2}
\end{figure}

An interesting case is Tottenham Hotspur, the 3rd-ranked team in the 2017/18 Premier League season. While they certainly played a remarkable season, both their offensive and defensive quality were on a high level. Figure \ref{fig:use_case2} allows us to investigate their season in more detail. Specifically, the figure considers the relative difference in predicted vs true possession values of teams. That is, for each possession, we compute the difference between the value for the possession according to our model and the actual ``observed'' possession value computed as described in Section \ref{sec:choosing_pv}. Then, we divide the difference by the predicted possession value to account for the fact that a higher predicted threat allows for a higher discrepancy from observed values. Finally, we aggregated the relative difference value for teams over the season, similar to what we did beforehand with the possession value. In this case, a smaller value (the value can potentially be negative) corresponds to a better performance. In general, we observe that the predicted value is higher than the actual value of the possession. Intuitively, this can be explained by the fact that the predicted value takes into account all possible options for the next action in each step of the possession. Hence, suboptimal decisions in terms of actions performed are accumulated within each possession.
A team with a smaller relative difference can be interpreted as being more effective in their attacking intent, as indicated by being closer to their potential possession value from the model. As in Figure \ref{fig:use_case1}, Panel A of Figure \ref{fig:use_case2} relates this difference in possession value to final points, whereas Panel B relates it to xG. In both cases, we observe a strong negative correlation ($R \approx -0.876$ in Panel A and $R \approx -0.907$ in Panel B), indicating that better teams generate more effective possessions, as shown by smaller accumulated relative difference in predicted vs true possession value. Panel C displays again the densities for this relative difference for the possessions of the six top-ranked teams. All three plots in Figure \ref{fig:use_case2} show that Tottenham played a very effective season, meaning that the relative difference between predicted and true possession value was generally low. This may be one aspect why they ended up in 3rd position overall.         

\subsection{Analyzing specific game situations}

\begin{figure}[t!]
    \centering
    \includegraphics[width=\textwidth]{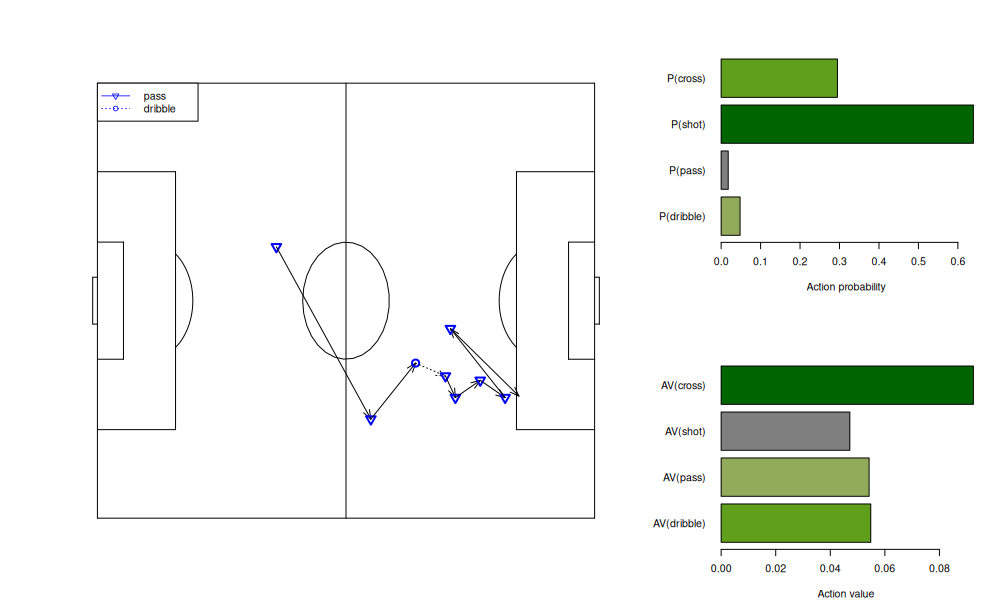}
    \caption{
    A visual representation of one possession in the data. The top barplot displays the action probabilities as predicted from our model. The bottom barplot displays the action value (LAV) when performing each of the possible actions. 
    }
    \label{fig:control_room}
\end{figure}

Beyond possession value analyses, our framework enables the evaluation of specific in-game situations in soccer, either in real time or as part of pre-game preparation. Figure \ref{fig:control_room} presents an example of how the model can be used to analyze a specific situation in detail. For the displayed possession, we are able to analyze the most likely next action as well as the value for certain options. In particular, our model expects a shot with high probability, but also assigns a fair portion of probability to a cross. For each action, we can additionally use our model to compute the hypothetical action value using \eqref{eq:AV} for performing any of the actions. It can be seen that the best option is to cross the ball instead of shooting the ball. Such a tool could support coaches in developing strategies and guiding players toward better decision-making. 

\section{Discussion and conclusion}

In this work, we present a novel methodology for predicting the next action in a soccer possession. We propose to use path signatures to encode the spatio-temporal information contained in possessions. Path signatures are a natural choice for the task of action predictions, because they are able to handle possessions of different lengths and they implicitly extract all relevant information in a time-series, avoiding the need for explicit feature engineering. Our results show that our model accurately predicts next actions and is computationally more efficient than popular existing methods. Furthermore, we present a novel method to evaluate possessions, taking into account action type and predicted locations in an intuitive and interpretable way. Our possession evaluation metric proves to be more reliable in predicting future performance than existing alternatives, and our action prediction model performs better paired with the metric than the benchmark model. We complement these findings with a detailed analysis of the 2017/18 Premier League season to show the effectiveness of our model and the possession evaluation metric. Finally, we present further use cases of our model in practice. 

There are various ways to extend our framework. For comparison reasons, we mainly followed the definitions of possessions and actions from previous work \citep{simpson2022seq2event,yeung2025transformer}. While we believe that this is a sensible approach, clearly capturing the most important aspects of soccer possessions, the data would allow for using a more granular set of action types. Depending on the interest of the analyst, our model could be implemented with a more diverse set of action types. Furthermore, for practical reasons, we refrained from explicitly modeling the interevent time. However, it is straightforward to adapt our framework to also account for the time between two events. 

Another key factor for improving and extending our work is the available data. For this paper, we used openly available event stream data from Wyscout. Our action prediction model and the resulting applications are limited to the information available within this data. More seasons of data, as well as more granular data such as tracking data, would allow for more refined analyses. In particular, signatures could be used in a similar fashion for tracking data time series to improve the use cases presented. To be more precise, if the trajectories of all 22 players within possessions were available, one could encode this information via signatures to obtain far more predictive models for the current state and value of a possession. Due to the spatio-temporal nature of sports data in general, we believe there is an immense potential for using path signatures in various settings to enhance sports analytics.

In conclusion, the framework presented in this work is a step toward better modeling and understanding professional soccer. Using path signatures as a tool for capturing the complex spatio-temporal structure of the sport enhances possession evaluation models and opens up novel possibilities for analyzing the inherently dynamic nature of many sports.

\backmatter

\bmhead{Supplementary information}

The code for reproducing the results and figures for this paper is available at \url{https://github.com/Rob2208/sig_actions}. Data preprocessing and implementation of the benchmark transformer model are based on code from \url{https://github.com/calvinyeungck/Football-Match-Event-Forecast/}, distributed under the Apache License 2.0. We modified the code to fit our experimental requirements. 

\begin{appendices}
\section{Hyperparameter Tuning}\label{appendix_hyperparameter}
We conducted an extensive hyperparameter tuning, systematically testing various parameter combinations. The values used during the experiments are stated in Table \ref{tab:params}, where the final selection is marked in bold.
\begin{table}[h]
\caption{Grid of evaluated hyperparameters, with final selection highlighted in bold.}\label{tab:params}%
\begin{tabular}{@{}ll@{}}
\toprule
Hyperparameter & Values\\
\midrule
CEL scaling    & \textbf{1}, 5   \\
Hidden dim    & 64, 128, \textbf{256}    \\
Batch size    & \textbf{4}, 10, 32  \\
Signature order    & \textbf{3}, 4   \\
\botrule
\end{tabular}
\end{table}
As mentioned in Section \ref{sec:tuning}, hyperparameters were selected in two phases based on model performance, using $n_r\in\{3,4,5,6,7\}$ recent action features as input, hence forecasting from the
4th, 5th, 6th, 7th or 8th action onward. In phase one, we investigate the CEL scaling parameter $\lambda$, signature orders $M$, and batch sizes $n_{batch}$. Table \ref{tab:cel} shows losses for the evaluated CEL scaling values, indicating the best overall performance for $\lambda=1$. 
\begin{table}[h]
    \centering
    \caption{Average losses across hyperparameter grid for CEL scaling $\lambda \in \{1,5\}$, for each value of historic actions $n_r\in\{3,4,5,6,7\}$.}\label{tab:cel}
    \begin{tabular}{c rr rr rr rr}
        \toprule
        \textbf{$n_r$}
        & \multicolumn{2}{c}{\textbf{MSE}}
        & \multicolumn{2}{c}{\textbf{CEL}}
        & \multicolumn{2}{c}{\textbf{Brier}}
        & \multicolumn{2}{c}{\textbf{KL}} \\[2pt]
        \cmidrule(lr){2-3}\cmidrule(lr){4-5}\cmidrule(lr){6-7}\cmidrule(lr){8-9}
        & $\lambda$=1 & $\lambda$=5
        & $\lambda$=1 & $\lambda$=5
        & $\lambda$=1 & $\lambda$=5
        & $\lambda$=1 & $\lambda$=5\\
        \midrule
        3 & 0.1643 & 0.1649 & 0.0488 & 0.0489 & 0.7985 & 0.8005 & 0.1122 & 0.1118 \\
        4 & 0.1648 & 0.1654 & 0.0496 & 0.0498 & 0.7962 & 0.7987 & 0.1143 & 0.1144 \\
        5 & 0.1648 & 0.1656 & 0.0507 & 0.0508 & 0.7899 & 0.7930 & 0.1159 & 0.1163 \\
        6 & 0.1656 & 0.1662 & 0.0513 & 0.0512 & 0.7893 & 0.7916 & 0.1174 & 0.1172 \\
        7 & 0.1655 & 0.1661 & 0.0523 & 0.0524 & 0.7927 & 0.7950 & 0.1183 & 0.1185 \\
        \bottomrule
    \end{tabular}
\end{table}

Table \ref{tab:sig} shows losses for the tested signature orders, indicating the best overall performance for $M=3$. 
\begin{table}[h]
    \centering
    \caption{Average losses across hyperparameter grid for signature order $M\in\{3,4\}$, for each value of historic actions $n_r\in\{3,4,5,6,7\}$.}\label{tab:sig}
    \begin{tabular}{c rr rr rr rr}
        \toprule
        \textbf{$n_r$}
        & \multicolumn{2}{c}{\textbf{MSE}}
        & \multicolumn{2}{c}{\textbf{CEL}}
        & \multicolumn{2}{c}{\textbf{Brier}}
        & \multicolumn{2}{c}{\textbf{KL}} \\[2pt]
        \cmidrule(lr){2-3}\cmidrule(lr){4-5}\cmidrule(lr){6-7}\cmidrule(lr){8-9}
        & M=3 & M=4
        & M=3 & M=4
        & M=3 & M=4
        & M=3 & M=4 \\
        \midrule
        3 & 0.1645 & 0.1642 & 0.0486 & 0.0489 & 0.7971 & 0.8000 & 0.1122 & 0.1122 \\
        4 & 0.1650 & 0.1646 & 0.0496 & 0.0497 & 0.7961 & 0.7963 & 0.1142 & 0.1144 \\
        5 & 0.1648 & 0.1648 & 0.0507 & 0.0508 & 0.7898 & 0.7901 & 0.1158 & 0.1161 \\
        6 & 0.1658 & 0.1655 & 0.0513 & 0.0513 & 0.7890 & 0.7896 & 0.1171 & 0.1178 \\
        7 & 0.1658 & 0.1653 & 0.0522 & 0.0523 & 0.7935 & 0.7919 & 0.1184 & 0.1182 \\
        \bottomrule
    \end{tabular}
\end{table}
In table \ref{tab:batch} we present the losses for our evaluated batch sizes, which led to the exclusion of batch size 32, due to its significantly higher RMSE. 

\begin{table}[h]
    \centering
    \caption{Average losses across hyperparameter grid for batch size $n_{\text{batch}}\in\{4,10,32\}$, for each value of historic actions $n_r\in\{3,4,5,6,7\}$.}
    \label{tab:batch}
\begin{tabular}{c rrr rrr}
        \toprule
        \textbf{$n_r$}
        & \multicolumn{3}{c}{\textbf{MSE}}
        & \multicolumn{3}{c}{\textbf{CEL}} \\[2pt]
        \cmidrule(lr){2-4}\cmidrule(lr){5-7}
        & 4 & 10 & 32
        & 4 & 10 & 32 \\
        \midrule
        3 & 0.1596 & 0.1657 & 0.1683 & 0.0486 & 0.0486 & 0.0486 \\
        4 & 0.1601 & 0.1662 & 0.1688 & 0.0497 & 0.0495 & 0.0496 \\
        5 & 0.1600 & 0.1657 & 0.1688 & 0.0507 & 0.0508 & 0.0506 \\
        6 & 0.1606 & 0.1670 & 0.1698 & 0.0512 & 0.0512 & 0.0515 \\
        7 & 0.1610 & 0.1666 & 0.1697 & 0.0519 & 0.0519 & 0.0529 \\
        \hline\\[-4pt] \addlinespace[5pt]
        \multicolumn{1}{c}{} & \multicolumn{3}{c}{\textbf{Brier}} & \multicolumn{3}{c}{\textbf{KL}} \\[2pt]
        \cmidrule(lr){2-4}\cmidrule(lr){5-7}
        & 4 & 10 & 32 & 4 & 10 & 32 \\
        3 & 0.7965 & 0.7965 & 0.7982 & 0.1119 & 0.1125 & 0.1122 \\
        4 & 0.7960 & 0.7956 & 0.7968 & 0.1135 & 0.1144 & 0.1147 \\
        5 & 0.7872 & 0.7910 & 0.7912 & 0.1151 & 0.1160 & 0.1163 \\
        6 & 0.7873 & 0.7887 & 0.7910 & 0.1164 & 0.1173 & 0.1176 \\
        7 & 0.7893 & 0.7946 & 0.7965 & 0.1170 & 0.1184 & 0.1199 \\
        \bottomrule
    \end{tabular}
\end{table}

During the second phase of hyperparameter tuning, we evaluated the six most promising model candidates. Their parameter configuration are shown in  Table \ref{tab:model_configs}.
\begin{table}[h]
    \centering
    \caption{Hyper-parameter settings for the six model candidates after tuning stage one.}
    \label{tab:model_configs}
    \begin{tabular}{c c c c l}
        \toprule
        \textbf{Model} & \textbf{CEL scale $\lambda$} & \textbf{signature order $M$} &
        \textbf{Batch size} & \textbf{Hidden dimension} \\
        \midrule
        model 1 & 1 & 3 & 4  & [64, 64]   \\
        model 2 & 1 & 3 & 4  & [128, 128] \\
        model 3 & 1 & 3 & 4  & [256, 256] \\
        model 4 & 1 & 3 & 10 & [64, 64]   \\
        model 5 & 1 & 3 & 10 & [128, 128] \\
        model 6 & 1 & 3 & 10 & [256, 256] \\
        \bottomrule
    \end{tabular}
\end{table}
We evaluate their robustness based on average losses across ten data sets, generated by randomly selecting teams and matches for the used train and test data sets. The average scores, along with their standard deviation across the generated datasets, are shown in Table \ref{tab:mean_std_all_models}.
\begin{table}[h!]
    \centering
    \caption{Mean (Std) across datasets for model candidates and Seq2Event (smallest value across model candidates per metric and number of historic actions in bold)
    Average losses (standard deviations) across test datasets for each model in tuning phase two. Best values for each metric are highlighted in bold, separately for each value of historic actions $n_r\in\{3,4,5,6,7\}$.
    }
    \label{tab:mean_std_all_models}
    \footnotesize
    \setlength{\tabcolsep}{3pt}
    \begin{tabular}{c c l l l l l}
        \toprule
        \textbf{$n_r$} & \textbf{Model} &
        \textbf{Test loss} & \textbf{MSE} & \textbf{CEL} &
        \textbf{KL} & \textbf{Brier} \\
        \midrule
\multirow{7}{*}{3}
  & 1 & 0.2091\,(0.0031) & 0.1601\,(0.0019) & 0.0491\,(0.0016) & 0.1126\,(0.0020) & 0.8033\,(0.0096) \\
  & 2 & 0.2092\,(0.0032) & 0.1600\,(0.0019) & 0.0492\,(0.0017) & 0.1126\,(0.0020) & 0.8028\,(0.0098) \\
  & 3 & \textbf{0.2090\,(0.0030)} & \textbf{0.1599\,(0.0018)} & \textbf{0.0490\,(0.0016)} & \textbf{0.1123\,(0.0019)} & 0.8022\,(0.0090) \\
  & 4 & 0.2151\,(0.0033) & 0.1661\,(0.0020) & \textbf{0.0490\,(0.0017)} & 0.1126\,(0.0021) & 0.8025\,(0.0091) \\
  & 5 & 0.2152\,(0.0032) & 0.1661\,(0.0018) & \textbf{0.0490\,(0.0017)} & 0.1128\,(0.0020) & 0.8020\,(0.0107) \\
  & 6 & 0.2151\,(0.0033) & 0.1661\,(0.0019) & \textbf{0.0490\,(0.0017)} & 0.1125\,(0.0020) & \textbf{0.8010\,(0.0096)} \\
        \midrule
\multirow{7}{*}{4}
  & 1 & 0.2104\,(0.0029) & 0.1603\,(0.0017) & 0.0501\,(0.0017) & 0.1146\,(0.0023) & 0.7973\,(0.0111) \\
  & 2 & 0.2104\,(0.0030) & \textbf{0.1601\,(0.0018)} & 0.0503\,(0.0018) & 0.1145\,(0.0021) & 0.7966\,(0.0107) \\
  & 3 & \textbf{0.2102\,(0.0031)} & \textbf{0.1601\,(0.0018)} & 0.0501\,(0.0018) & \textbf{0.1140\,(0.0023)} & 0.7947\,(0.0103) \\
  & 4 & 0.2166\,(0.0031) & 0.1666\,(0.0018) & \textbf{0.0500\,(0.0017)} & 0.1147\,(0.0023) & 0.7955\,(0.0110) \\
  & 5 & 0.2164\,(0.0031) & 0.1664\,(0.0019) & \textbf{0.0500\,(0.0017)} & 0.1145\,(0.0025) & 0.7953\,(0.0112) \\
  & 6 & 0.2164\,(0.0032) & 0.1662\,(0.0019) & 0.0501\,(0.0018) & 0.1141\,(0.0024) & \textbf{0.7946\,(0.0105)} \\
        \midrule
\multirow{7}{*}{5}
  & 1 & \textbf{0.2112\,(0.0035)} & 0.1604\,(0.0020) & \textbf{0.0509\,(0.0020)} & 0.1156\,(0.0026) & 0.7906\,(0.0109) \\
  & 2 & \textbf{0.2112\,(0.0035)} & \textbf{0.1602\,(0.0019)} & 0.0510\,(0.0020) & 0.1156\,(0.0030) & 0.7907\,(0.0107) \\
  & 3 & \textbf{0.2112\,(0.0033)} & \textbf{0.1602\,(0.0017)} & 0.0510\,(0.0021) & \textbf{0.1154\,(0.0028)} & 0.7913\,(0.0107) \\
  & 4 & 0.2176\,(0.0036) & 0.1667\,(0.0019) & \textbf{0.0509\,(0.0021)} & 0.1162\,(0.0034) & 0.7922\,(0.0134) \\
  & 5 & 0.2176\,(0.0037) & 0.1667\,(0.0021) & 0.0510\,(0.0021) & 0.1158\,(0.0029) & \textbf{0.7901\,(0.0112)} \\
  & 6 & 0.2173\,(0.0035) & 0.1664\,(0.0019) & \textbf{0.0509\,(0.0020)} & 0.1156\,(0.0026) & 0.7906\,(0.0106) \\
        \midrule
\multirow{7}{*}{6}
  & 1 & 0.2127\,(0.0035) & 0.1610\,(0.0018) & 0.0517\,(0.0022) & 0.1172\,(0.0024) & 0.7934\,(0.0129) \\
  & 2 & \textbf{0.2122\,(0.0038)} & \textbf{0.1605\,(0.0019)} & 0.0516\,(0.0023) & 0.1165\,(0.0025) & 0.7906\,(0.0130) \\
  & 3 & 0.2124\,(0.0039) & 0.1606\,(0.0020) & 0.0518\,(0.0022) & 0.1166\,(0.0025) & \textbf{0.7904\,(0.0146)} \\
  & 4 & 0.2186\,(0.0038) & 0.1671\,(0.0020) & \textbf{0.0515\,(0.0022)} & 0.1170\,(0.0024) & 0.7918\,(0.0129) \\
  & 5 & 0.2187\,(0.0039) & 0.1671\,(0.0021) & 0.0517\,(0.0023) & 0.1169\,(0.0024) & 0.7906\,(0.0129) \\
  & 6 & 0.2186\,(0.0035) & 0.1670\,(0.0019) & 0.0517\,(0.0021) & \textbf{0.1164\,(0.0026)} & 0.7914\,(0.0130) \\
        \midrule
\multirow{7}{*}{7}
  & 1 & \textbf{0.2141\,(0.0041)} & 0.1618\,(0.0020) & \textbf{0.0523\,(0.0025)} & 0.1170\,(0.0029) & 0.7936\,(0.0156) \\
  & 2 & 0.2143\,(0.0038) & \textbf{0.1616\,(0.0020)} & 0.0527\,(0.0024) & 0.1176\,(0.0031) & 0.7952\,(0.0161) \\
  & 3 & 0.2145\,(0.0037) & 0.1617\,(0.0018) & 0.0528\,(0.0024) & \textbf{0.1164\,(0.0030)} & \textbf{0.7917\,(0.0162)} \\
  & 4 & 0.2197\,(0.0039) & 0.1674\,(0.0019) & \textbf{0.0523\,(0.0023)} & 0.1176\,(0.0035) & 0.7955\,(0.0166) \\
  & 5 & 0.2199\,(0.0041) & 0.1674\,(0.0020) & 0.0525\,(0.0025) & 0.1177\,(0.0030) & 0.7953\,(0.0139) \\
  & 6 & 0.2196\,(0.0039) & 0.1672\,(0.0020) & 0.0524\,(0.0023) & 0.1175\,(0.0032) & 0.7952\,(0.0145) \\
        \bottomrule
    \end{tabular}
\end{table}

\section{Additional tests}\label{additional_tests}
To support our pre-processing proposed in \ref{sec:data_pp}, we evaluate our model both with and without the additional hand-crafted features proposed in \cite{simpson2022seq2event}, with their results presented in Table \ref{tab:with_additional_features}.
\begin{table}[h]
    \centering
    \caption{Losses of our Sig-Model with and without handcrafted features (HC) proposed in \cite{simpson2022seq2event}, where Test loss = MSE + CEL. Best values for each $n_r\in\{3,4,5,6,7\}$ are highlighted in bold.}
    \label{tab:with_additional_features}
    \begin{tabular}{c c r r r r r}
        \toprule
        \textbf{$n_r$} & \textbf{Sig-Model} & \textbf{Test loss} & \textbf{MSE} & \textbf{CEL} & \textbf{Brier} & \textbf{KL} \\
        \midrule
        \multirow{2}{*}{3}
            & w/o HC   & \textbf{0.2084} & \textbf{0.1598} & 0.0486 & \textbf{0.7968} & \textbf{0.1117} \\
            & w/ HC  & 0.2092 & 0.0507 & 0.1584 & 0.8034 & 0.1133 \\
        \midrule
        \multirow{2}{*}{4}
            & w/o HC   & \textbf{0.2096} & \textbf{0.1601} & 0.0495 & \textbf{0.7938} & \textbf{0.1125} \\
            & w/ HC   & 0.2105 & 0.0515 & 0.1590 & 0.8008 & 0.1138  \\
        \midrule
        \multirow{2}{*}{5}
            & w/o HC   & \textbf{0.2108} & \textbf{0.1601} & 0.0507 & \textbf{0.7877} & \textbf{0.1146} \\
            & w/ HC   & 0.2111 & 0.0526 & 0.1586 & 0.7951 & 0.1163 \\
        \midrule
        \multirow{2}{*}{6}
            & w/o HC   & \textbf{0.2121} & \textbf{0.1606} & 0.0515 & \textbf{0.7858} & \textbf{0.1155} \\
            & w/ HC &  0.2123 & 0.0526 & 0.1597 & 0.7893 & 0.1167 \\
        \midrule
        \multirow{2}{*}{7}
            & w/o HC   & \textbf{0.2134} & \textbf{0.1612} & 0.0522 & \textbf{0.7881} & \textbf{0.1169} \\
            & w/ HC &  0.2160 & 0.0560 & 0.1600 & 0.7945 & 0.1201 \\
        \bottomrule
    \end{tabular}
\end{table}
Moreover, we adapt the code from \url{https://github.com/calvinyeungck/Football-Match-Event-Forecast/} and re-implemented the Seq2Event model proposed by \cite{simpson2022seq2event} for various configurations and report its results in \ref{tab:loss_comparison}. We varied the length of past actions history used as model input during training or restricted the input to the same raw match data triplets $(x,y,T)$ used in our approach without engineering additional geometric or temporal features. This analysis provides empirical evidence for the effectiveness of our proposed pre-processing and the reduced dependency of our approach on hand-crafted features. It further highlights the ability of path signatures to capture rich spatial-temporal structures directly from raw data.
\begin{table}[h!]
    \centering
    \caption{Losses of Sig2Event model for three different historic window sizes $n_r\in\{5,10,40\}$ and for window size 40 only using $(x,y,T)$ as input for each $n_r\in\{3,4,5,6,7\}$. Test loss is the sum of MSE and CEL).}
    \label{tab:transformer_losses}
    \begin{tabular}{c l r r r r r}
        \toprule
        \textbf{$n_r$} & \textbf{Specification} &
        \textbf{Test loss} & \textbf{MSE} & \textbf{CEL} &
        \textbf{KL} & \textbf{Brier} \\
        \midrule
\multirow{4}{*}{3}
  & 40 past actions  & 0.4036 & 0.1652 & 0.2383 & 0.1121 & 0.8034 \\
  & 10 past actions  & 0.3994 & 0.1644 & 0.2350 & 0.1140 & 0.8066 \\
  & 5  past actions  & 0.4014 & 0.1644 & 0.2370 & 0.1138 & 0.8053 \\
  & $x,y,T$ (40 past actions)          & 0.4051 & 0.1666 & 0.2385 & 0.1144 & 0.8086 \\\midrule
\multirow{4}{*}{4}
  & 40 past actions  & 0.4087 & 0.1658 & 0.2429 & 0.1163 & 0.8054 \\
  & 10 past actions  & 0.4047 & 0.1645 & 0.2402 & 0.1168 & 0.8033 \\
  & 5  past actions  & 0.4039 & 0.1638 & 0.2401 & 0.1166 & 0.8050 \\
  & $x,y,T$ (40 past actions)         & 0.4100 & 0.1661 & 0.2439 & 0.1151 & 0.8029 \\\midrule
\multirow{4}{*}{5}
  & 40 past actions  & 0.4101 & 0.1652 & 0.2449 & 0.1177 & 0.7984 \\
  & 10 past actions  & 0.4111 & 0.1651 & 0.2460 & 0.1164 & 0.7962 \\
  & 5  past actions  & 0.4079 & 0.1639 & 0.2440 & 0.1219 & 0.7995 \\
  & $x,y,T$ (40 past actions)         & 0.4128 & 0.1658 & 0.2470 & 0.1197 & 0.7979 \\\midrule
\multirow{4}{*}{6}
  & 40 past actions  & 0.4120 & 0.1646 & 0.2474 & 0.1183 & 0.7954 \\
  & 10 past actions  & 0.4090 & 0.1637 & 0.2453 & 0.1207 & 0.7972 \\
  & 5  past actions  & 0.4087 & 0.1633 & 0.2454 & 0.1187 & 0.7948 \\
  & $x,y,T$ (40 past actions)         & 0.4159 & 0.1681 & 0.2478 & 0.1186 & 0.7925 \\\midrule
\multirow{4}{*}{7}
  & 40 past actions  & 0.4193 & 0.1642 & 0.2551 & 0.1202 & 0.7998 \\
  & 10 past actions  & 0.4164 & 0.1637 & 0.2527 & 0.1211 & 0.8023 \\
  & 5  past actions  & 0.4160 & 0.1635 & 0.2524 & 0.1224 & 0.8041 \\
  & $x,y,T$ (40 past actions)         & 0.4209 & 0.1647 & 0.2562 & 0.1194 & 0.8005 \\
        \bottomrule
    \end{tabular}
\end{table}

\section{Possession value details}\label{app:pvd}

Expected goals (xG) models have emerged as popular tools for a more granular analysis of team and player performance. These models assign a probability of success to each shot, taking into account factors that influence the likelihood of scoring a goal from a shot. The earliest version of an xG model has already been proposed by \cite{PR97xG}. The authors used a logistic regression to model the binary shot data and found that the most important factors for successful shots were the shot location and the angle between the shot and the two goalposts (henceforth, goal angle). Due to advancements in data collection and computing, nowadays powerful machine learning models are employed for the task of fitting an xG model \cite{Robberechts20xG,Anzer21xG}. 

For this work and our possession value metric, we are only interested in a proxy for shot quality based on location and therefore follow a simple approach based on \cite{PR97xG} to fit an xG model. In particular, we model binary shots data via a logistic regression model, i.e., a model assuming a linear relationship between the log-odds of the probability of scoring from a shot $\pi(Z) = P(Y = 1 \mid Z)$ and features $Z$:
\begin{equation}\label{eq:xG_lrm}
\log\left(\frac{\pi(Z)}{1-\pi(Z)}\right) = Z^{\top}\gamma,
\end{equation}
where the feature vector $Z$ contains the distance to the goal and the goal angle. We train our xG models on data from four big European leagues (Bundesliga, La Liga, Ligue 1, and Serie A), but we do not use Premier League data, to not interfere with the data used for our analyses in Section \ref{sec:practice}.

Expected threat (xT) was first popularized through a blog \citep{Singh19xT}. The idea is to estimate a value of each game state, where a game state is dependent only on the location on the pitch, via a Markov model. In particular, the pitch is divided into a grid of size $M \times N$ resulting in $M \cdot N$ location zones, for each of which a game state value is derived. These values for zone $z$ are derived by iteratively solving the equation
\begin{equation}\label{eq:xT}
\operatorname{xT}(z) = P(S = 1|z) \cdot \operatorname{xG} + (1-P(S = 1|z)) \sum_{i = 1}^{M\cdot N} T_{z,i} \operatorname{xT}(i),
\end{equation}
where $P(S=1|z)$ is the probability that a player shoots at location $z$ (hence, $1-P(S = 1|z)$ is the probability of moving the ball instead of shooting), $\operatorname{xG}$ is a simple xG model as above, and $T_{z,i}$ is the $(z,i)$-th element of a transition matrix $T$, i.e., represents the probability of transitioning from zone $z$ to zone $i$. To solve the problem, the values for xT are initialized to zero and iterated for a number of times (ideally until some convergence is achieved). Intuitively, an xT value for zone $z$ derived from $k$ iterations represents the probability of scoring from the next $k$ actions, when starting in zone $z$. We fitted the above model on the same data as before with a $12 \times 16$ grid. 




\end{appendices}


\bibliography{bib_ML}

\end{document}